\documentclass{article}

\PassOptionsToPackage{numbers, compress}{natbib}
\usepackage[main,final]{neurips_2026}

\usepackage[utf8]{inputenc}
\usepackage[T1]{fontenc}
\usepackage{hyperref}
\usepackage{url}
\usepackage{booktabs}
\usepackage{amsfonts}
\usepackage{amssymb}
\usepackage{amsmath}
\usepackage{nicefrac}
\usepackage{microtype}
\usepackage[table]{xcolor}
\usepackage{graphicx}
\usepackage{multirow}
\usepackage{makecell}
\usepackage{threeparttable}
\usepackage{enumitem}
\usepackage[most]{tcolorbox}
\usepackage{listings}
\usepackage{subcaption}
\usepackage{bm}
\usepackage{tikz}
\usepackage{colortbl}
\usepackage{algorithm}
\usepackage{algorithmic}
\usepackage{wrapfig}

\graphicspath{{./}}

\title{REFLEX: Reflective Evolution from LLM Experience}

\author{
  Pan Wang \\
  University of Science and Technology of China \\
  \texttt{diogenescask@gmail.com}
}

\begin{document}

\maketitle

\begin{abstract}
Large multimodal language models (MLLMs) have emerged as powerful tools for guiding evolutionary search toward interpretable programmatic policies. In existing program-evolution systems, however, reusable knowledge is usually carried by whole programs in the population, and it is difficult to trace how a visual observation led to a particular code change and its measured outcome. We present REFLEX, a train-free evolutionary framework that links these steps in one loop. A vision-enabled Critic turns task-specific behavioral evidence into a structured diagnosis; the diagnosis retrieves executable code snippets from a persistent Skill Memory; a text-only Actor writes the child program; and the child--parent fitness change updates the utility of each retrieved snippet. Every step is recorded in a single trace. Under matched backends and 100-call budgets over 10 paired seeds, REFLEX reaches the solve threshold in a median of 13, 20, and 24 LLM calls on Acrobot, Pendulum, and Lunar Lander, roughly half the calls required by official MLES and by a compute-matched Actor-only ablation. Frozen Skill Memory banks from Pendulum or Lunar Lander raise the final Acrobot score on all 10 paired seeds. On a 36-dimensional antenna-array design task with equal evaluation budgets and shared initialization, REFLEX reaches the harder $25.25$ score threshold on 9/10 seeds, compared with at most 3/10 for CMA-ES, GA, and PSO. The source code is available at https://github.com/wangpan-ustc/REFLEX.
\end{abstract}

\section{Introduction}
\label{sec:intro}

Deep reinforcement learning (DRL) has achieved remarkable success in complex control tasks~\cite{luo2024end}, yet its resulting policies are typically black-box neural networks that are notoriously difficult to interpret, verify, or manually patch~\cite{lin2024hierarchical,shindo2024blendrl,vincze2025smose,xiao2025joint}. Unlike black-box approaches, programmatic policies express controllers as human-readable source code~\cite{verma2018programmatically}. This paradigm retains the expressiveness of modern programming while ensuring the safety and auditability necessary for real-world applications~\cite{liu2024synthesizing,trivedi2021learning,kohler2024interpretable,yuan2026sp,xiao2026yang,xiao2025optical,xiao2025robotic,xiao2026opening}. Recently, large language models (LLMs) have emerged as powerful evolutionary operators capable of discovering such policies~\citep{hu2026seal,wang2026diagnosis,kuang2025learning,nie2026understanding}. Multimodal Large Language Model-assisted Evolutionary Search (MLES)~\citep{hu2026mles} shows that multimodal behavioral evidence from policy executions can be leveraged by MLLMs to diagnose failure patterns and guide evolutionary program search toward more effective programmatic controllers~\citep{wang2026atlasva,zhu2026medeyes}.

However, in existing frameworks like MLES, each BE-conditioned mutation is still a single multimodal call that must simultaneously interpret complex visual trajectories, diagnose the underlying failure, and synthesize the corrective code~\cite{liu2024evolution,madaan2023self,shinn2023reflexion,yao2023tree}. This \textit{diagnosis--repair entanglement} creates an opaque feedback loop. When a proposed mutation fails, it is difficult to audit whether the model misinterpreted the visual evidence or simply generated flawed code~\cite{xia2025demystifying,xu2025aligning,tong2024can,tong2026ask,zhu2026craft}. Furthermore, reusable knowledge in these systems is mostly kept inside individual programs rather than as executable code whose usefulness is measured~\cite{tong2026grounded}, so algorithmic insights and code heuristics discovered in previous trials are hard to carry into new campaigns~\cite{jiang2024ledex,yang2024swe}.

To address these fundamental bottlenecks, we introduce \textbf{REFLEX} (\underline{R}eflective \underline{E}volution \underline{F}rom \underline{L}LM \underline{Ex}perience), a fully train-free evolutionary framework that, following the reflection--generation separation of prior systems such as Reflexion~\cite{shinn2023reflexion} and ReEvo~\cite{ye2024reevo}, decouples visual diagnosis from code generation. In REFLEX, a specialized vision-enabled \textit{Critic} first analyzes the Behavioral Evidence and emits a structured, traceable JSON diagnosis outlining specific failure modes and root causes. Subsequently, a text-optimized \textit{Actor} reads this diagnosis alongside the parent code to synthesize a targeted repair. Crucially, REFLEX augments the Actor with a persistent, self-evolving \textit{Skill Memory}---a library of reusable code snippets extracted from elite individuals and credited with the fitness change of the children that use them. This enables the framework to abstract, store, and transfer learned programmatic concepts (e.g., energy-pumping heuristics or altitude-gated braking) across independent evolutionary campaigns.

We evaluate REFLEX across a diverse suite of domains, encompassing both classic control benchmarks (Lunar Lander, Acrobot, Pendulum) and a highly non-convex 36-dimensional engineering optimization problem (compact Concentric Circular Antenna Array synthesis) \cite{dolph1946,dib2014ccaa}. Under matched 100-call budgets over 10 paired seeds, REFLEX reaches the solve threshold in a median of $13$, $20$, and $24$ LLM calls on Acrobot, Pendulum, and Lunar Lander, compared with $41$, $53$, and $49$ for official MLES and $28$, $46$, and $43$ for a compute-matched Actor-only ablation. It also achieves a final Lunar Lander NWS of $1.082 \pm 0.009$, and in the antenna design task reaches a score of $25.25$ on 9/10 seeds under equal evaluation budgets and shared initialization, compared with at most 3/10 for CMA-ES, GA, and PSO.

In summary, our main contributions are:
\begin{itemize}
    \item We identify the diagnosis--repair entanglement bottleneck in multimodal LLM-guided evolution and propose REFLEX, which adopts a decoupled Critic-Actor design from prior reflective systems~\cite{shinn2023reflexion,ye2024reevo} to make programmatic mutations traceable and targeted.
    \item We introduce a self-evolving Skill Memory mechanism that credits reusable code snippets with the fitness change they produce and enables cross-run programmatic knowledge transfer across evolutionary campaigns.
    \item We demonstrate that, under matched budgets, REFLEX achieves strong sample efficiency across discrete and continuous control environments and generalizes to a real-world engineering design task.
\end{itemize}

\section{Related Work}
\label{sec:related}

\paragraph{LLM-Guided Evolutionary Search.}
LLMs are now widely used as mutation and crossover operators in evolutionary algorithms \cite{lehman2023evolution, romera2024mathematical, liu2024evolution,guo2023connecting,todd2024gavel,van2024llamea}. FunSearch \cite{romera2024mathematical} evolves programs against an automated evaluator and keeps them in a program database; EoH \cite{liu2024evolution} co-evolves natural-language thoughts and code; LLaMEA \cite{van2024llamea} generates and refines complete metaheuristics; ReEvo \cite{ye2024reevo} adds short- and long-term verbal reflections that steer subsequent code generation; and AlphaEvolve \cite{novikov2025alphaevolve} and its open-source counterpart OpenEvolve \cite{sharma2025openevolve} combine several LLMs, evaluators, and a program database. MLES \cite{hu2026mles} brings visual Behavioral Evidence into this setting, with multiple BE-conditioned operators for programmatic policy search. In all of these systems the evolved object is a program and the main store of reusable knowledge is the population or program database, possibly accompanied by textual reflections. REFLEX shares this backbone and, like ReEvo and AlphaEvolve, uses distinct LLM roles, but also stores snippet-level executable code that is retrieved by the visual diagnosis and credited with the child--parent fitness change it produces.

\paragraph{Programmatic Policies and Agent Memory.}
Programmatic policies offer a transparent and verifiable alternative to black-box neural networks in deep reinforcement learning (DRL) \cite{verma2018programmatically,verma2019imitation,trivedi2021learning}. Eureka \cite{ma2023eureka} uses LLMs to write reward functions for a DRL inner loop, whereas REFLEX evolves the executable policy directly. Self-improving agents increasingly rely on persistent memory~\cite{zhu2026path,zhu2026medsynapse}. Voyager \cite{wang2023voyager} stores executable skills retrieved by natural-language description and verified by self-checking; Agentic Context Engineering \cite{zhang2026ace} incrementally curates a textual playbook of strategies; ReasoningBank \cite{ouyang2026reasoningbank} distills reasoning memories from successful and failed trajectories; and EvoTest \cite{he2026evotest} evolves an entire agent configuration (prompt, memory, hyperparameters, and tool routines) between episodes. REFLEX stores executable snippets, retrieves them with BM25 queries formed from a visual diagnosis, updates their utility from measured child--parent fitness within a population-based search, and is evaluated for cross-run and cross-task transfer with frozen banks~\cite{xiong2026dive}.

\paragraph{Self-Reflection and Visual Diagnostics.}
Self-reflection and critique improve LLM reasoning and coding~\cite{shinn2023reflexion,chen2023teaching,zhang2023self,gou2023critic,weng2023large,xiong2026enhancing,xiong2025deliberate,lin2026models}, and Trace \cite{cheng2024trace} optimizes computational workflows from execution traces and rich feedback. 
Recent work has further examined the reliability of LLM-based evaluators, distinguishing systematic between-prompt variation from stochastic within-prompt sampling noise~\cite{feng2026second}. Most reflective frameworks consume text: execution logs, compiler errors, or traces~\cite{bouzenia2025repairagent,zhang2024autocoderover,zhong2024debug}. In continuous control and physical design, scalar rewards and text logs often miss the geometric or spatial character of a failure. Like MLES, REFLEX reads rendered behavioral evidence; its vision-language Critic writes a structured diagnosis that is logged and reused as the retrieval query. We apply the same loop to classic control and to Concentric Circular Antenna Array (CCAA) synthesis \cite{dib2014ccaa,dolph1946}, an engineering design problem usually addressed with black-box heuristics (e.g., GA, PSO) or analytical tapers.

\section{Method}
\label{sec:method}

\subsection{Problem Formulation and Overview}
\begin{figure}[t]
  \centering
  \includegraphics[width=\linewidth]{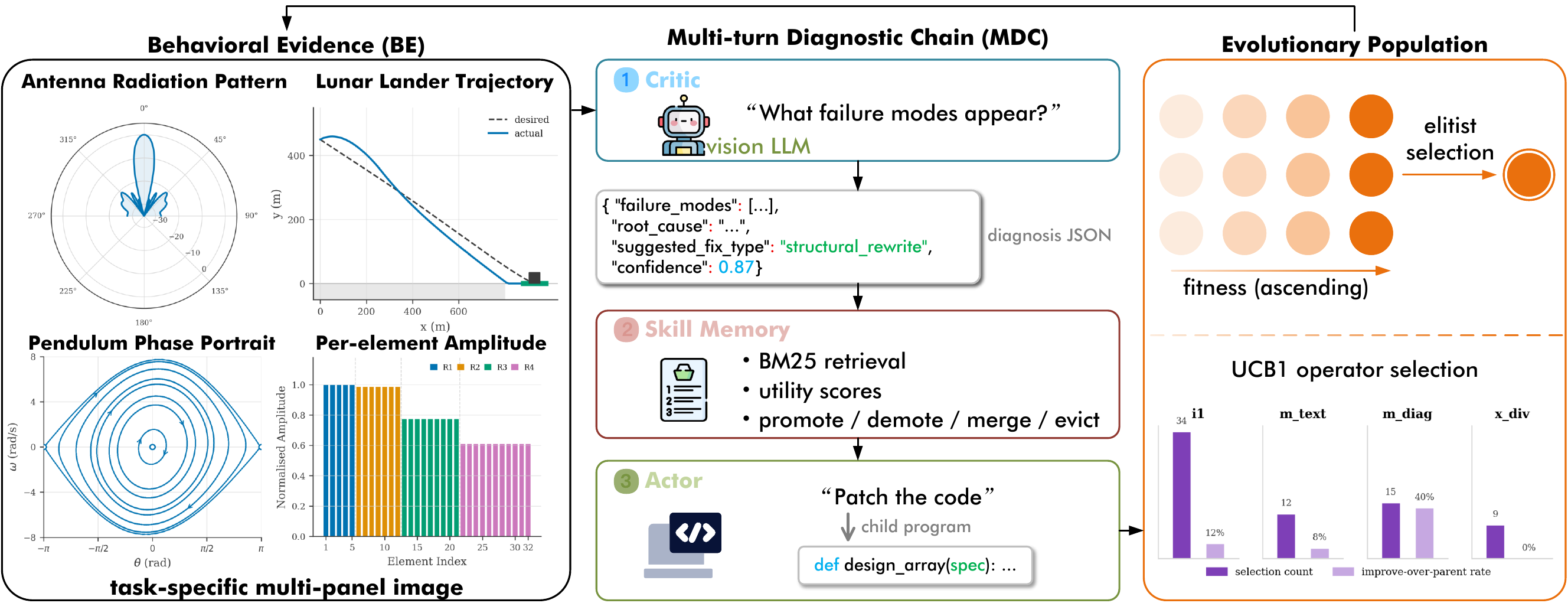}
  \caption{Overview of the REFLEX framework. The evolutionary loop is driven by a Multi-turn Diagnostic Chain (MDC) that decouples visual diagnosis from code generation. First, task-specific Behavioral Evidence (BE) is analyzed by a vision-enabled Critic to produce a structured JSON diagnosis. Next, a text-optimized Actor synthesizes a child program by combining the parent code, the Critic's diagnosis, and reusable code snippets retrieved from a persistent Skill Memory. The generated programs are evaluated and maintained in an evolutionary population using elitist selection and UCB1 operator selection.}
  \label{fig:overview}
\end{figure}

We model programmatic policy search as the optimization of a human-readable source code program $p \in \mathcal{P}$ to maximize a task-specific fitness function $\mathcal{F}(p)$. Evaluating a candidate policy $p$ produces both a scalar fitness score $f_p = \mathcal{F}(p)$ and a visual Behavioral Evidence (BE) image $V_p = \mathcal{B}(p)$. To solve $p^* = \arg\max_{p \in \mathcal{P}} \mathcal{F}(p)$, REFLEX employs an evolutionary framework driven by large language models. As illustrated in Figure~\ref{fig:overview}, REFLEX consists of three core modules: (1) a Multi-turn Diagnostic Chain (MDC) that decouples visual diagnosis from code generation, (2) task-specific Behavioral Evidence, and (3) a persistent Skill Memory. A UCB1 bandit selects among evolutionary operators. We write $\Pi_t \subset \mathcal{P}$ for the current finite population.

\subsection{Critic-Actor Decoupling via Multi-turn Diagnostic Chain}
When a single generation step both interprets the image and writes code, the record of a mutation does not show which of the two went wrong. Following prior reflective systems~\cite{shinn2023reflexion,ye2024reevo}, the Multi-turn Diagnostic Chain (MDC) assigns these steps to separate calls whose inputs and outputs are logged. 

Given a parent policy $p$, its fitness $f_p$, and its BE image $V_p$, a vision-enabled Critic first analyzes the visual evidence. The Critic is modeled as a conditional distribution $P_\text{critic}(D_p \mid V_p, f_p, \mathcal{T})$, where $\mathcal{T}$ is the task description and $D_p$ is a structured JSON diagnosis containing failure modes and root causes. The Critic does not write code. 

Subsequently, a text-optimized Actor synthesizes a child program $p'$. The Actor samples from $P_\text{actor}(p' \mid p, D_p, S_p, \mathcal{T})$, where $S_p$ represents relevant code snippets retrieved from the Skill Memory (detailed in Section~\ref{sec:skill_memory}). The Actor never sees the raw image $V_p$. 

Each mutation records the diagnosed visual problem $D_p$, the retrieved skills $S_p$, and the child $p'$ with its measured fitness, so a failed mutation can be traced to the diagnosis or to the code edit.

\begin{algorithm}[!h]
\caption{REFLEX: Reflective Evolution from LLM Experience (as run in the control tasks)}
\label{alg:reflex}
\begin{algorithmic}[1]
\STATE \textbf{Input:} task $\mathcal{T}$, fitness $\mathcal{F}$, BE renderer $\mathcal{B}$, template program $p_0$, population capacity $N$, operators $\mathcal{O}$, stagnation threshold $s_{\max}$, LLM-call budget $B$ (every Critic, Actor, repair, revision, abstraction, merge, and failed call counts)
\STATE Evaluate $p_0$ and cache $(f_{p_0}, V_{p_0})$; $\Pi_t \leftarrow \{p_0\}$; $\mathcal{M} \leftarrow \mathcal{M}_0$ (empty or a frozen bank); $N_o, \hat{\mu}_o \leftarrow 0 \ \forall o \in \mathcal{O}$; $s \leftarrow 0$
\STATE Fill $\Pi_t$ to $N/2$ members with diagnostic or text-only mutations of $p_0$
\WHILE{fewer than $B$ LLM calls have been attempted}
    \IF{$s \ge s_{\max}$}
        \STATE $Q \leftarrow (\texttt{m\_inspire}, \texttt{i1})$; $s \leftarrow 0$ \COMMENT{forced burst}
    \ELSE
        \STATE $Q \leftarrow (o_t)$ with $o_t \in \arg\max_{o \in \mathcal{O}} U_t(o)$ (Eq.~\ref{eq:ucb})
    \ENDIF
    \FOR{each operator $o$ in $Q$}
        \STATE $p \leftarrow \textsc{Tournament}_3(\Pi_t)$; $D_p \leftarrow \varnothing$; $S_p \leftarrow \emptyset$
        \IF{$o = \texttt{m\_diag}$}
            \STATE $D_p \sim P_\text{critic}(\cdot \mid V_p, f_p, \mathcal{T})$ \COMMENT{reused for at most two Actor calls on $p$}
            \STATE $S_p \leftarrow$ top-$K$ BM25 skills for $D_p$ in $\mathcal{M}$; $p' \sim P_\text{actor}(\cdot \mid p, D_p, S_p, \mathcal{T})$
        \ELSIF{$o = \texttt{x\_div}$}
            \STATE $p' \sim P_\text{actor}(\cdot \mid \text{three mutually diverse members of } \Pi_t, \mathcal{T})$
        \ELSE
            \STATE $p' \sim P_\text{actor}(\cdot \mid p, f_p, \mathcal{T})$ for \texttt{m\_text}; the same plus a foreign-domain mechanism for \texttt{m\_inspire}; a new program shown the top-3 skills by $u_i$ for \texttt{i1}
        \ENDIF
        \STATE Evaluate $p'$ and cache $(f_{p'}, V_{p'})$; if it crashes, make one repair call and re-evaluate
        \STATE If $p$ is near the best and $p'$ regressed, one revision call yields a second child, processed like $p'$ below
        \STATE $\Delta \leftarrow f_{p'} - f_p$, or $\Delta \leftarrow -1$ if $p'$ is invalid
        \STATE Update $u_i$ of the skills in $S_p$ that the Actor reports using, then apply the archive rules (Sec.~\ref{sec:skill_memory})
        \STATE \textbf{if} $\Delta > 0$ \textbf{then} abstract one new skill from $p'$ (one LLM call) and add it to $\mathcal{M}$
        \STATE Update $N_o, \hat{\mu}_o$ with $r = \min(\max(\Delta, 0), \Delta_{\max}) / \Delta_{\max}$; if $r > 0$, credit ancestors' operators with $0.5^d r$
        \STATE Insert $p'$ into $\Pi_t$ if valid and not a duplicate, then keep the $N$ best
        \STATE $s \leftarrow 0$ if $f_{p'}$ is a new best, else $s \leftarrow s + 1$
    \ENDFOR
\ENDWHILE
\STATE \textbf{Return:} $p^* = \arg\max_{p \in \Pi_t} f_p$
\end{algorithmic}
\end{algorithm}

\subsection{Behavioral Evidence}
Scalar rewards $f_p$ often hide the specific reasons for failure (e.g., late braking in Lunar Lander or high sidelobes in antenna synthesis). We design BE to visually expose these failure modes.

BE is a task-specific rendering function $V_p = \mathcal{B}(p)$. For control environments, $V_p$ summarizes multi-step trajectories, phase portraits, and action timelines. For Concentric Circular Antenna Array (CCAA) synthesis, $V_p$ visualizes the element layout, amplitude taper, and array-factor cuts. 

By deliberately designing $V_p$ to be diagnostic, the Critic can reliably ground its JSON diagnosis $D_p$ in observable physical or geometric anomalies rather than guessing from a single scalar reward.

\subsection{Skill Memory and Evolutionary Operators}
\label{sec:skill_memory}
In population-based LLM-guided evolution, a useful heuristic usually persists only inside the programs that contain it. REFLEX additionally stores such heuristics as executable snippets with a fitness-grounded utility, and uses an operator bandit to allocate mutations.

REFLEX maintains a Skill Memory $\mathcal{M} = \{ (k_i, c_i, u_i) \}_{i=1}^{|\mathcal{M}|}$, where each skill comprises a natural language trigger $k_i$, a code snippet $c_i$, and a utility score $u_i$. During the MDC, the Critic's diagnosis $D_p$ acts as a query to retrieve the top-$K$ skills $S_p \subset \mathcal{M}$ using the BM25 scoring function:
\begin{equation}
\text{score}(D_p, k_i) = \sum_{q \in D_p} \text{IDF}(q) \cdot \frac{f(q, k_i) \cdot (k_1 + 1)}{f(q, k_i) + k_1 \cdot \left(1 - b + b \cdot \frac{|k_i|}{\text{avgdl}}\right)}
\end{equation}
where $f(q, k_i)$ is the term frequency of query term $q$ in trigger $k_i$. Retrieval ranks skills by this score alone. After a child $p'$ of parent $p$ is evaluated, each retrieved skill that the Actor reports using is credited with $\Delta = f_{p'} - f_p$ ($\Delta = -1$ for an invalid child): $u_i \leftarrow u_i + \min(\Delta, 1)$ if $\Delta > 0$, and $u_i \leftarrow u_i - \tfrac{1}{2}\min(|\Delta|, 1)$ otherwise. New skills start at $u_i = 1$. A skill is archived when $u_i < -1$ after at least three retrievals, near-duplicate skills are merged by one LLM call, and when the bank exceeds its capacity of 256 active skills the lowest-utility skills are archived. Besides these archive rules, $u_i$ is shown to the Actor next to each retrieved snippet, selects the skills shown to \texttt{i1}, and selects the frozen banks used for transfer.

New skills are extracted only when a child improves on its parent ($\Delta > 0$). In such cases, one LLM call abstracts the child's code, its concept, and the diagnosis it addressed into a short snippet $c_i$ and a one-sentence trigger $k_i$.

To manage the evolutionary loop, REFLEX uses five operators, which form $\mathcal{O}$ in this order: diagnostic mutation through the MDC (\texttt{m\_diag}); text-only mutation from the parent's code and score (\texttt{m\_text}); crossover of three mutually diverse population members (\texttt{x\_div}); generation of a new program from scratch (\texttt{i1}); and cross-domain inspiration (\texttt{m\_inspire}), which asks the Actor to transfer a mechanism from a randomly drawn unrelated domain into the parent. We model operator selection as a multi-armed bandit solved with UCB1. At step $t$, the framework selects $o_t \in \arg\max_{o \in \mathcal{O}} U_t(o)$, where
\begin{equation}
U_t(o) = \hat{\mu}_o + C \sqrt{\frac{\ln t}{N_o}}
\label{eq:ucb}
\end{equation}
$\hat{\mu}_o$ is the mean normalized reward of operator $o$, $N_o$ is the number of times $o$ has been selected, and $C$ balances exploration and exploitation. Untried operators ($N_o = 0$) are selected first, and ties are broken by the order of $\mathcal{O}$ listed above. To satisfy the bounded-reward assumption of UCB1, the fitness delta $\Delta$ is clipped to $[0, \Delta_{\max}]$ and divided by $\Delta_{\max}$, the largest positive delta observed so far in the run. In the control tasks, a positive reward $r$ is also credited, as $0.5^d r$, to the operators that produced the child's ancestors at depths $d \le 3$, and after $12$ consecutive children without a new best, REFLEX runs a forced burst of one \texttt{m\_inspire} and one \texttt{i1} proposal.

\begin{figure}[t]
  \centering
  \begin{subfigure}[b]{0.48\linewidth}
    \centering
    \includegraphics[width=\linewidth]{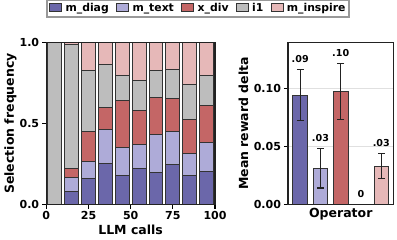}
    \caption{Operator selection frequency and reward delta.}
    \label{fig:operator_bandit}
  \end{subfigure}
  \hfill
  \begin{subfigure}[b]{0.48\linewidth}
    \centering
    \includegraphics[width=\linewidth]{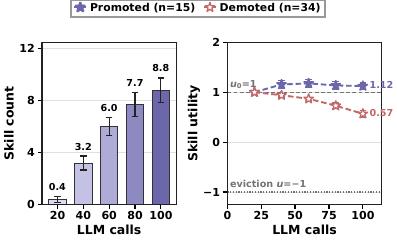}
    \caption{Skill memory size and utility tracking.}
    \label{fig:skill_memory}
  \end{subfigure}
  \caption{Evolutionary dynamics in REFLEX. (a) The UCB1 bandit dynamically balances diagnostic mutation (\texttt{m\_diag}), text mutation (\texttt{m\_text}), crossover (\texttt{x\_div}), cross-domain inspiration (\texttt{m\_inspire}), and initialization (\texttt{i1}) based on their historical success. (b) The Skill Memory continuously evaluates extracted code snippets, promoting skills whose children improve on their parents, demoting the others, and archiving skills whose utility falls below $-1$.}
  \label{fig:operator_and_skill}
\end{figure}

As illustrated in Figure~\ref{fig:operator_and_skill}, the UCB1 bandit dynamically adapts to the search phase (Figure~\ref{fig:operator_bandit}), while the Skill Memory continuously evaluates extracted code snippets (Figure~\ref{fig:skill_memory}). Frozen snapshots of the Skill Memory can be loaded into new campaigns, which we use for the controlled transfer study in Section~\ref{sec:transfer}. The complete evolutionary loop, integrating the Critic's visual diagnosis, Actor's synthesis, and Skill Memory updates, is formalized in Algorithm~\ref{alg:reflex}.

\section{Experiments}
\label{sec:experiments}
\subsection{Experimental Setup}
\textbf{Tasks and Metrics.} We evaluate REFLEX on four tasks. Lunar Lander uses a Normalized Weighted Score (NWS) combining reward, fuel use, and success rate; Pendulum uses a stable-upright score; and Acrobot reports the raw evaluator score (higher is better, target $-0.20$). The fourth task is Concentric Circular Antenna Array (CCAA) synthesis, a 36-dimensional continuous design problem optimizing element amplitudes and ring radii for a 5-7-9-11 compact array; its score rewards lower sidelobe level (SLL), directivity above 14 dB, and closeness to a 28-degree first-null beamwidth target. For the control tasks we report the final best score, the AUC of the best-so-far curve over the call budget, the number of seeds reaching the task target (Solved), and the number of LLM calls needed to first reach it (calls to solve).

\textbf{Baselines and Protocol.} We compare REFLEX with two baselines under matched conditions: (i) official MLES \citep{hu2026mles}, rerun from its public implementation with its native prompts, operators, parent and environmental selection, and duplicate handling; and (ii) Actor-only, a compute-matched ablation that removes the Critic and spends the saved calls on additional Actor proposals. Its \texttt{m\_diag} operator keeps its bandit arm, but the Actor receives a placeholder diagnosis that lists only the parent's scalar evaluation metrics, and this text also serves as the BM25 query; no BE image reaches any LLM call, and all other operators and Skill Memory rules are unchanged. All methods use the same model weights (Qwen3-VL-235B-A22B-Instruct) for every LLM role, population size 16, a 4,096-token output cap, identical evaluators. The budget counts every LLM generation attempt, including Critic, Actor, repair, abstraction, and failed calls. The 50- and 100-call budgets are separate runs with different random seeds; within each budget, comparisons are paired by seed: we report 95\% paired-bootstrap confidence intervals, exact Wilcoxon signed-rank tests with Holm correction within each planned family, and Cohen's $d_z$. The cross-task transfer study (Section~\ref{sec:transfer}) uses Qwen3.5-27B. All hyperparameters and protocol details are listed in Appendix~\ref{sec:supp_protocol}.

\subsection{Sample-Efficient Policy Discovery}
\begin{figure}[htbp]
  \centering
  \includegraphics[width=\linewidth]{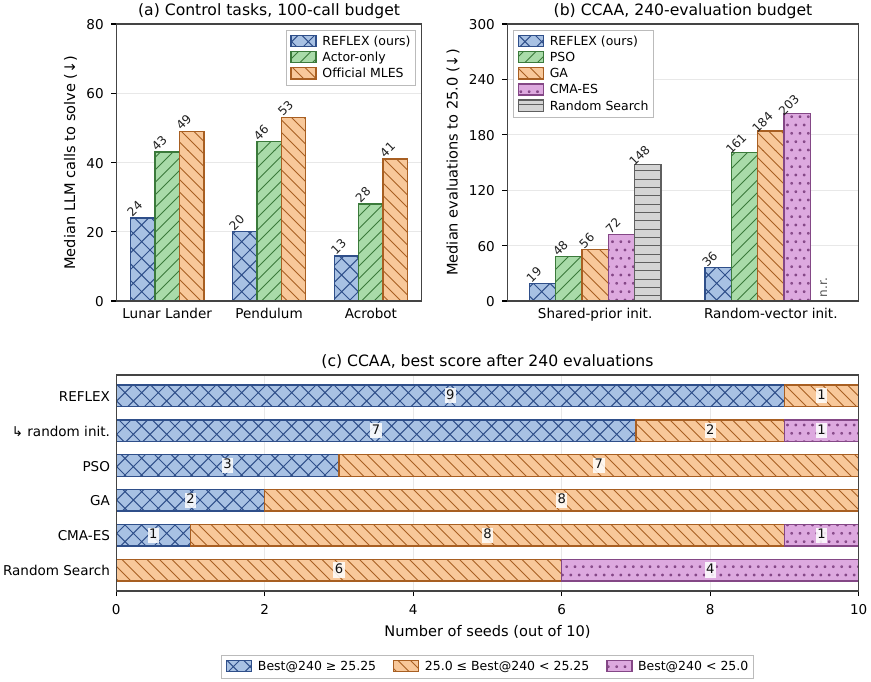}
  \caption{Sample efficiency under matched budgets (10 paired seeds; lower is better in (a) and (b)). \textbf{(a)} Median LLM calls to solve on the control tasks with the same backend and a cap of 100 total LLM calls, where every Critic, Actor, repair, and failed attempt counts. \textbf{(b)} Median objective evaluations to reach a CCAA score of $25.0$ under an equal 240-evaluation budget, with shared-prior or random-vector initialization; n.r.\ marks a method that never reaches $25.0$. \textbf{(c)} Number of CCAA seeds whose Best@240 falls in each score band. All rows use shared-prior initialization except ``$\hookrightarrow$ random init.'', which denotes REFLEX with random-vector initialization.}
  \label{fig:convergence}
\end{figure}

Table~\ref{tab:main_results} reports the matched comparison, and Figure~\ref{fig:convergence}(a) summarizes the median calls to solve at the 100-call budget. At 50 calls, REFLEX has the highest final score, AUC, and solve count on all three tasks. On Lunar Lander and Pendulum, its paired AUC gains over both baselines are significant at both budgets (Holm-corrected $p \le 0.032$); for example, on Pendulum at 50 calls the AUC gain is $+0.125$ [$0.045$, $0.205$] over Actor-only ($d_z = 0.90$) and $+0.160$ [$0.070$, $0.250$] over MLES ($d_z = 1.10$). Final-score gains have 95\% CIs excluding zero on Pendulum at both budgets and on Lunar Lander at 50 calls (e.g., $+0.037$ [$0.008$, $0.066$] over Actor-only), but not on Lunar Lander at 100 calls, where all three methods reach a similar final NWS. On Acrobot (without paired tests), REFLEX solves every seed at both budgets with a median of 12--13 calls, versus 24--28 for Actor-only and 35--41 for MLES.

REFLEX spends about 40\% of its calls on the Critic and therefore evaluates fewer policies: $27$ on average in a 50-call Lunar Lander campaign, versus $48$ for Actor-only and $42$ for MLES. It also produces fewer invalid programs; on Acrobot at 50 calls, 2 of 360 REFLEX children (0.6\%) are invalid, versus 75 of 510 (14.7\%) for MLES. With a 20-call cap on Acrobot, REFLEX and Actor-only each solve 6/10 seeds and their paired differences are not significant (Actor-only minus REFLEX: final $+0.015$, 95\% CI [$-0.123$, $0.453$], $p = 0.92$; AUC $+0.020$, $p = 0.63$); the Critic calls pay off once the budget allows several diagnosis-conditioned mutations.

\begin{table}[htbp]
  \centering
  \caption{Matched comparison under 50 and 100 total LLM calls (same model weights for all roles, population 16, 10 paired seeds; every LLM attempt counts). Final and AUC are mean $\pm$ SEM, except in the Acrobot 50-call block, where Final is the median ($^\dagger$) and AUC is the mean without SEM. Solved counts seeds that reach the task target; Calls is the median [IQR] number of LLM calls needed to first reach it. Lunar Lander reports NWS, Pendulum the stable-upright score, and Acrobot the raw evaluator score (target $-0.20$). Best value in each budget block is in \textbf{bold}.}
  \label{tab:main_results}
  \renewcommand{\arraystretch}{1.15}
  \setlength{\tabcolsep}{3pt}
  \resizebox{\linewidth}{!}{%
  \begin{tabular}{ll cccc cccc cccc}
    \toprule
    & & \multicolumn{4}{c}{\textbf{Lunar Lander (NWS)}}
      & \multicolumn{4}{c}{\textbf{Pendulum}}
      & \multicolumn{4}{c}{\textbf{Acrobot (raw)}} \\
    \cmidrule(lr){3-6} \cmidrule(lr){7-10} \cmidrule(lr){11-14}
    \textbf{Budget} & \textbf{Method}
      & Final$\uparrow$ & AUC$\uparrow$ & Solved & Calls$\downarrow$
      & Final$\uparrow$ & AUC$\uparrow$ & Solved & Calls$\downarrow$
      & Final$\uparrow$ & AUC$\uparrow$ & Solved & Calls$\downarrow$ \\
    \midrule
    \multirow{3}{*}{50}
      & Official MLES
      & $1.010_{\pm 0.020}$ & $0.860_{\pm 0.028}$ & 6/10 & 41 {\scriptsize[32,47]}
      & $0.735_{\pm 0.050}$ & $0.575_{\pm 0.046}$ & 6/10 & 39 {\scriptsize[29,48]}
      & $-0.145^\dagger$ & $-0.500$ & 9/10 & 35 {\scriptsize[23,47]} \\
      & Actor-only
      & $1.018_{\pm 0.018}$ & $0.895_{\pm 0.024}$ & 7/10 & 35 {\scriptsize[27,43]}
      & $0.770_{\pm 0.045}$ & $0.610_{\pm 0.042}$ & 6/10 & 34 {\scriptsize[25,45]}
      & $-0.148^\dagger$ & $-0.460$ & 9/10 & 24 {\scriptsize[15,36]} \\
      & \textbf{REFLEX}
      & $\mathbf{1.055}_{\pm 0.012}$ & $\mathbf{0.965}_{\pm 0.018}$ & \textbf{9/10} & \textbf{22} {\scriptsize[16,29]}
      & $\mathbf{0.875}_{\pm 0.030}$ & $\mathbf{0.735}_{\pm 0.035}$ & \textbf{9/10} & \textbf{18} {\scriptsize[12,27]}
      & $\mathbf{-0.135}^\dagger$ & $\mathbf{-0.320}$ & \textbf{10/10} & \textbf{12} {\scriptsize[7,20]} \\
    \midrule
    \multirow{3}{*}{100}
      & Official MLES
      & $1.069_{\pm 0.011}$ & $0.955_{\pm 0.018}$ & 9/10 & 49 {\scriptsize[36,66]}
      & $0.860_{\pm 0.034}$ & $0.715_{\pm 0.036}$ & 8/10 & 53 {\scriptsize[38,72]}
      & $-0.165_{\pm 0.015}$ & $-0.490_{\pm 0.055}$ & 9/10 & 41 {\scriptsize[27,61]} \\
      & Actor-only
      & $1.064_{\pm 0.012}$ & $0.970_{\pm 0.017}$ & 9/10 & 43 {\scriptsize[32,58]}
      & $0.875_{\pm 0.030}$ & $0.750_{\pm 0.032}$ & 9/10 & 46 {\scriptsize[31,63]}
      & $-0.155_{\pm 0.012}$ & $-0.410_{\pm 0.050}$ & \textbf{10/10} & 28 {\scriptsize[17,43]} \\
      & \textbf{REFLEX}
      & $\mathbf{1.082}_{\pm 0.009}$ & $\mathbf{1.025}_{\pm 0.012}$ & \textbf{10/10} & \textbf{24} {\scriptsize[17,31]}
      & $\mathbf{0.935}_{\pm 0.020}$ & $\mathbf{0.845}_{\pm 0.025}$ & \textbf{10/10} & \textbf{20} {\scriptsize[13,29]}
      & $\mathbf{-0.145}_{\pm 0.008}$ & $\mathbf{-0.280}_{\pm 0.035}$ & \textbf{10/10} & \textbf{13} {\scriptsize[7,21]} \\
    \bottomrule
  \end{tabular}%
  }
\end{table}

\subsection{Generalization to Engineering Design: CCAA Synthesis}
To answer how far the method can generalize under harder, out-of-distribution settings, we apply REFLEX to the CCAA synthesis task. Classical taper families such as Dolph-Chebyshev provide strong starting points for low-sidelobe arrays \citep{dolph1946}, and modern CCAA synthesis studies optimize radii, rotations, and excitations to further reduce sidelobes \citep{dib2014ccaa}. 

REFLEX uses the same evolutionary interface as the control tasks: the policy code emits a design vector, the evaluator computes SLL, FNBW, and directivity, and the BE exposes the array layout and cuts.

Because the LLM already knows common array designs, we give REFLEX, Random Search, CMA-ES, GA, and PSO the same initialization on the same evaluator: either random design vectors, or a shared prior consisting of the same first LLM/analytical designs for every method. All methods receive the same budget of 240 objective evaluations, and every evaluation counts, including REFLEX's inner local-search evaluations. Table~\ref{tab:ccaa_baselines} and Figure~\ref{fig:convergence}(b,c) report 10 paired seeds. The shared prior helps every classical optimizer substantially (e.g., GA reaches $25.0$ on 10/10 seeds instead of 4/10). Under both initializations REFLEX is the most sample-efficient method: with the shared prior it reaches $25.0$ in a median of 19 evaluations, versus 48--148 for the baselines, and attains a Best@240 of $25.33 \pm 0.02$. The gap is largest at the harder $25.25$ threshold, which REFLEX reaches on 9/10 seeds with the shared prior and 7/10 from random vectors, while no classical optimizer exceeds 3/10 in either setting.

\begin{table}[htbp]
  \centering
  \small
  \caption{CCAA synthesis under an equal budget of 240 objective evaluations (10 paired seeds; all evaluations count, including REFLEX's inner local search). Best@$k$ is the best score after $k$ evaluations (mean $\pm$ SEM). The threshold columns count seeds whose Best@240 reaches $25.0$ or $25.25$; Evals-to-25.0 is the median [IQR] over seeds that reach it (--: never reached).}
  \label{tab:ccaa_baselines}
  \resizebox{\linewidth}{!}{%
  \begin{tabular}{llcccccc}
    \toprule
    \textbf{Method} & \textbf{Initialization} & \textbf{Best@50}$\uparrow$ & \textbf{Best@100}$\uparrow$ & \textbf{Best@240}$\uparrow$ & $\boldsymbol{\ge 25.0}$ & $\boldsymbol{\ge 25.25}$ & \textbf{Evals-to-25.0}$\downarrow$ \\
    \midrule
    Random Search & Random vector & $23.82 \pm 0.18$ & $24.10 \pm 0.15$ & $24.48 \pm 0.12$ & 0/10 & 0/10 & -- \\
    CMA-ES & Random vector & $24.28 \pm 0.13$ & $24.55 \pm 0.11$ & $24.83 \pm 0.09$ & 2/10 & 0/10 & 203 [175, 231] \\
    GA & Random vector & $24.34 \pm 0.12$ & $24.66 \pm 0.10$ & $24.95 \pm 0.07$ & 4/10 & 0/10 & 184 [136, 220] \\
    PSO & Random vector & $24.40 \pm 0.11$ & $24.72 \pm 0.09$ & $25.00 \pm 0.06$ & 5/10 & 0/10 & 161 [119, 206] \\
    \textbf{REFLEX} & Random vector & $\mathbf{24.86 \pm 0.06}$ & $\mathbf{25.11 \pm 0.04}$ & $\mathbf{25.28 \pm 0.03}$ & \textbf{9/10} & \textbf{7/10} & \textbf{36} [22, 57] \\
    \midrule
    Random Search & Shared prior & $24.86 \pm 0.07$ & $24.94 \pm 0.06$ & $25.03 \pm 0.05$ & 6/10 & 0/10 & 148 [104, 211] \\
    CMA-ES & Shared prior & $24.94 \pm 0.05$ & $25.04 \pm 0.04$ & $25.12 \pm 0.03$ & 9/10 & 1/10 & 72 [48, 119] \\
    GA & Shared prior & $24.96 \pm 0.04$ & $25.07 \pm 0.03$ & $25.18 \pm 0.03$ & \textbf{10/10} & 2/10 & 56 [37, 91] \\
    PSO & Shared prior & $24.98 \pm 0.04$ & $25.10 \pm 0.03$ & $25.21 \pm 0.02$ & \textbf{10/10} & 3/10 & 48 [31, 78] \\
    \textbf{REFLEX} & Shared prior & $\mathbf{25.02 \pm 0.04}$ & $\mathbf{25.18 \pm 0.03}$ & $\mathbf{25.33 \pm 0.02}$ & \textbf{10/10} & \textbf{9/10} & \textbf{19} [11, 31] \\
    \bottomrule
  \end{tabular}}
\end{table}

\begin{figure}[htbp]
  \centering
  \includegraphics[width=0.8\linewidth]{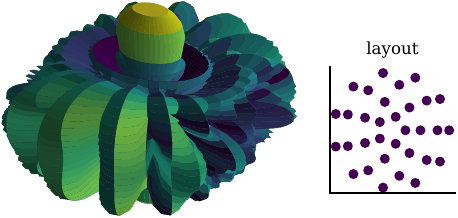}
  \caption{Best compact CCAA radiation pattern discovered in the existing REFLEX campaign.}
  \label{fig:antenna3d}
\end{figure}

\begin{figure}[htbp]
  \centering
  \begin{subfigure}[b]{0.58\linewidth}
    \centering
    \includegraphics[width=\linewidth]{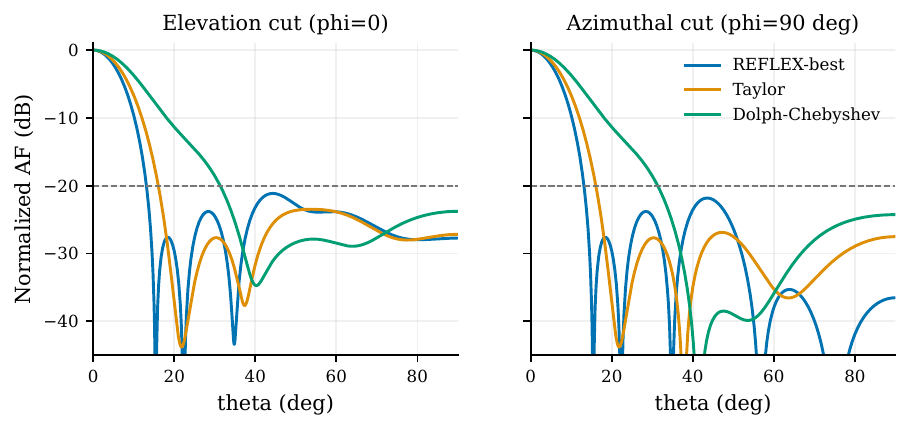}
    \caption{Elevation and azimuthal cuts.}
    \label{fig:antenna_cuts}
  \end{subfigure}
  \hfill
  \begin{subfigure}[b]{0.38\linewidth}
    \centering
    \includegraphics[width=\linewidth]{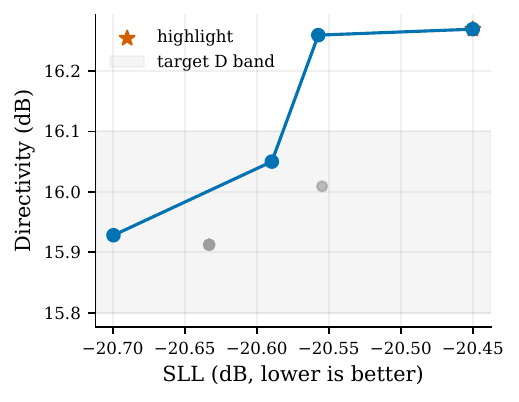}
    \caption{SLL vs. Directivity Pareto front.}
    \label{fig:pareto}
  \end{subfigure}
  \caption{Detailed analysis of the CCAA synthesis results. (a) Radiation pattern cuts comparing the REFLEX-best design against classical Dolph-Chebyshev and Taylor tapers, showing superior sidelobe suppression. (b) The Pareto front of discovered designs, illustrating the trade-off between Sidelobe Level (SLL) and Directivity within the target band.}
  \label{fig:antenna_analysis}
\end{figure}

Figure~\ref{fig:antenna_analysis} provides a detailed analysis of the optimized array. The elevation and azimuthal cuts (Figure~\ref{fig:antenna_cuts}) demonstrate that the REFLEX-best design achieves lower sidelobe levels compared to classical Dolph-Chebyshev and Taylor tapers. Furthermore, the Pareto front (Figure~\ref{fig:pareto}) illustrates the trade-off between sidelobe level and directivity explored during the evolutionary search, highlighting the target design band. Figure~\ref{fig:antenna3d} visualizes the resulting 3D radiation pattern.

\subsection{Controlled Skill Memory Transfer}
\label{sec:transfer}
We test whether executable Skill Memory transfers across tasks with a controlled three-condition study on Acrobot: an empty bank (\textit{cold}), the top-2 skills from Pendulum campaigns, and the top-2 skills from Lunar Lander campaigns. Both source banks are frozen and hash-verified before the target runs, selected by the same source-task utility rule, and audited to exclude Acrobot-specific identifiers. Each condition uses the same model (Qwen3.5-27B), a 20-call budget, and 10 paired campaign seeds; all 30 runs and 600 model attempts passed validation.

\begin{figure}[!h]
  \centering
  \begin{minipage}[t]{0.48\linewidth}
    \centering
    \includegraphics[width=\linewidth]{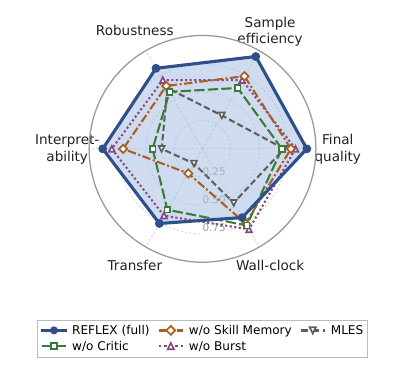}
    \caption{Qualitative component profile across sample efficiency, quality, robustness, interpretability, transfer, and wall-clock time axes.}
    \label{fig:ablation}
  \end{minipage}
  \hfill
  \begin{minipage}[t]{0.48\linewidth}
    \centering
    \includegraphics[width=\linewidth]{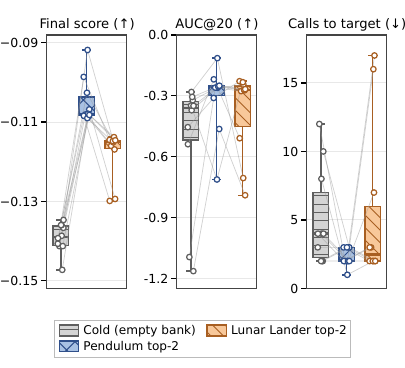}
    \caption{Cross-task Skill Memory transfer from Pendulum and Lunar Lander to Acrobot (10 paired seeds, 20-call budget).}
    \label{fig:warm_cold}
  \end{minipage}
\end{figure}

In Figure~\ref{fig:warm_cold}, boxes show medians and IQRs, whiskers span min--max, and linked dots indicate paired seeds across conditions. Both source banks raise the final score on every seed (10/10 paired wins each; exact Wilcoxon $p=0.002$), from a cold mean of $-0.139$ to $-0.105$ (Pendulum) and $-0.118$ (Lunar Lander). AUC@20 improves on 9/10 seeds for each bank, with means increasing from $-0.523$ to $-0.310$ and $-0.380$, respectively, although the paired test is significant only for the Pendulum bank ($p=0.049$; Lunar Lander $p=0.065$). Median calls-to-target fall from $4.0$ to $2.0$ and $2.5$, but this speedup is consistent only for the Pendulum bank (6/2/2 W/T/L, $p=0.039$); with the Lunar Lander bank two seeds need 16--17 calls ($p=0.91$). The trace shows that transfer works through the Actor rewriting retrieved code rather than copying it: Pendulum skills were used in 58 evaluated Acrobot children (13 improvements, 45 regressions) and Lunar Lander skills in 14 (5 improvements, 9 regressions). Fitness selection keeps the few useful adaptations, so transfer improves whole campaigns even though most individual reuses cause regressions.

\subsection{Ablation Studies}
\label{sec:ablation}
The compute-matched Actor-only ablation on all three control tasks is part of Table~\ref{tab:main_results}. Table~\ref{tab:ablation} adds two Acrobot ablations with matched call accounting and 10 paired seeds.

\begin{table}[t]
  \centering
  \small
  \caption{Acrobot ablations (10 paired seeds; Target counts seeds reaching $-0.20$). \textbf{(a)} Behavioral Evidence given to the Critic, under an identical call budget for all conditions; values are mean $\pm$ SEM. \textbf{(b)} Object stored in a frozen source Skill Memory, with a 20-call cap; Final is the median and AUC@20 the mean over seeds.}
  \label{tab:ablation}
  \begin{tabular}{lcccc}
    \toprule
    \multicolumn{5}{l}{\textit{(a) Behavioral Evidence perturbation}} \\
    \textbf{Condition} & \textbf{Final}$\uparrow$ & \textbf{AUC}$\uparrow$ & \textbf{Target} & \textbf{Calls-to-target}$\downarrow$ \\
    \midrule
    Full REFLEX (correct BE) & $\mathbf{-0.1352 \pm 0.0041}$ & $\mathbf{-0.2850 \pm 0.0337}$ & 10/10 & $\mathbf{5.50 \pm 1.58}$ \\
    Shuffled-BE Critic & $-0.1426 \pm 0.0040$ & $-0.4270 \pm 0.0556$ & 10/10 & $14.70 \pm 4.37$ \\
    Blank-BE Critic & $-0.1388 \pm 0.0020$ & $-0.3620 \pm 0.0408$ & 10/10 & $9.50 \pm 2.55$ \\
    Actor-only (no Critic) & $-0.1373 \pm 0.0021$ & $-0.3450 \pm 0.0347$ & 10/10 & $10.30 \pm 3.34$ \\
    \midrule
    \multicolumn{5}{l}{\textit{(b) Memory object}} \\
    \textbf{Memory} & \textbf{Final}$\uparrow$ & \textbf{AUC@20}$\uparrow$ & \textbf{Target} & \textbf{Invalid children}$\downarrow$ \\
    \midrule
    Executable fragments & $\mathbf{-0.1363}$ & $-0.5445$ & 9/10 & \textbf{0/163} \\
    Textual strategies & $-0.1393$ & $\mathbf{-0.4742}$ & \textbf{10/10} & 1/166 \\
    No memory & $-0.1417$ & $-0.5618$ & 9/10 & 3/160 \\
    \bottomrule
  \end{tabular}
\end{table}

\noindent\textbf{Behavioral Evidence.} To separate the effect of the visual evidence from that of an extra Critic call, we keep the Critic but give it a shuffled BE image that does not belong to the parent, or a blank image. Correct BE gives the best AUC and the fastest target attainment (Table~\ref{tab:ablation}a). Its paired AUC gain is $+0.142$ [$0.095$, $0.181$] over shuffled BE ($p = 0.004$), $+0.077$ [$0.040$, $0.110$] over blank BE ($p = 0.012$), and $+0.060$ [$0.027$, $0.089$] over Actor-only ($p = 0.012$), and it needs $9.2$, $4.0$, and $4.8$ fewer calls to reach the target, respectively. The final-score gain over Actor-only is not significant ($+0.0021$ [$-0.0005$, $0.0046$], $p = 0.18$).

\noindent\textbf{Memory object.} With the same frozen source bank and a 20-call cap, executable fragments, textual strategies, and no memory lead to different search profiles (Table~\ref{tab:ablation}b), but no pairwise difference is significant (textual minus executable, median of the paired differences: final $-0.0010$ [$-0.0093$, $0.0087$], $p = 0.82$; AUC $+0.017$ [$-0.194$, $0.265$], $p = 0.70$). Executable fragments produce no invalid children (0/163).

Figure~\ref{fig:ablation} provides a qualitative component profile of the Lunar Lander variants, in which the full REFLEX framework scores highest on sample efficiency, final quality, robustness, and interpretability.

\section{Conclusion and Limitations}
\label{sec:limitations}
REFLEX links visual Behavioral Evidence, a structured diagnosis, retrieved executable skills, the generated child, and its measured fitness in one loop, and credits each retrieved snippet with the fitness change it produces. With matched backends and total-call budgets of 50--100 calls, it reaches solve thresholds in roughly half the calls of official MLES and of a compute-matched Actor-only ablation. Shuffling or blanking the Behavioral Evidence removes this gain, and frozen Skill Memory banks from two source tasks improve search on Acrobot. The same loop also applies to a 36-dimensional antenna design task.

\textbf{Limitations.} At a 20-call cap on Acrobot, the Critic brings no measurable benefit over Actor-only. Our control tasks are low-dimensional and CCAA is a single structured design problem; building informative BE for high-dimensional observations or long-horizon control remains open. We have not run a human study of whether individual diagnoses are correct, nor a sensitivity study of population size, the UCB1 constant, or memory capacity. Cross-task transfer is tested on two source--target pairs with one model, and executable memory does not outperform textual memory in our ablation. REFLEX also inherits the stochasticity and cost of LLM inference.

\section*{Acknowledgments and Disclosures}

This work received no third-party funding or support.
The authors declare no competing interests.

\clearpage
\bibliographystyle{plainnat}
\bibliography{custom}

\clearpage
\appendix
\section{Prompt Templates and Examples}
\label{sec:supp_prompts}

To facilitate reproducibility, we provide the core prompt templates used by the Critic and Actor modules in the REFLEX framework. We use \texttt{tcolorbox} to present these prompts clearly.

\begin{tcolorbox}[
    enhanced,
    colback=red!5!white,
    colframe=red!60!black,
    boxrule=1pt,
    arc=4pt,
    title=Critic Prompt - Visual Diagnosis,
    fonttitle=\bfseries,
    attach boxed title to top left={xshift=10pt, yshift=-2mm},
    boxed title style={colback=red!60!black, colframe=red!60!black, arc=2pt}
]
\small
\textbf{System Role:} You are an expert control systems engineer and visual diagnostician.

\textbf{Input:}
1. Task Description: \{task\_description\}
2. Parent Policy Code:
```python
\{parent\_code\}
```
3. Fitness Score: \{fitness\_score\}
4. Behavioral Evidence Image: [Attached Image]

\textbf{Instruction:}
Analyze the provided Behavioral Evidence image. This image visualizes the execution of the parent policy. Your goal is to identify why the policy failed to achieve a perfect score. 

Do NOT write any code. Output your analysis STRICTLY as a JSON object with the following keys:
- \texttt{failure\_mode}: A short (2-4 words) categorization of the error.
- \texttt{visual\_evidence}: What specific physical/geometric anomalies do you see in the image?
- \texttt{root\_cause}: What algorithmic flaw in the parent code likely caused this visual behavior?
- \texttt{suggested\_fix}: How should the code be modified to fix this?

\textbf{Output Format:}
```json
\{
  "failure\_mode": "...",
  "visual\_evidence": "...",
  "root\_cause": "...",
  "suggested\_fix": "..."
\}
```
\end{tcolorbox}

\vspace{1em}

\begin{tcolorbox}[
    enhanced,
    colback=blue!5!white,
    colframe=blue!60!black,
    boxrule=1pt,
    arc=4pt,
    title=Actor Prompt - Code Synthesis,
    fonttitle=\bfseries,
    attach boxed title to top left={xshift=10pt, yshift=-2mm},
    boxed title style={colback=blue!60!black, colframe=blue!60!black, arc=2pt}
]
\small
\textbf{System Role:} You are an expert Python programmer specializing in control algorithms.

\textbf{Input:}
1. Task Description: \{task\_description\}
2. Parent Policy Code:
```python
\{parent\_code\}
```
3. Critic's Diagnosis:
```json
\{critic\_diagnosis\}
```
4. Retrieved Skills (from Skill Memory):
\{retrieved\_skills\}

\textbf{Instruction:}
Your task is to write a new, improved child policy. 
1. Read the Critic's diagnosis carefully to understand the failure mode and the suggested fix.
2. Review the retrieved skills. If any skill snippet is relevant to the suggested fix (e.g., an energy-pumping heuristic), adapt and integrate it into your new code.
3. Write the complete, executable Python function for the new policy.

\textbf{Output Format:}
Provide ONLY the Python code block. Do not include any conversational text.

\begin{verbatim}
def choose_action(state):
    # Your improved implementation here
    ...
\end{verbatim}
\end{tcolorbox}

\section{Interpretability of Critic Diagnoses}
\label{sec:supp_interpretability}

A central claim of the REFLEX framework is that decoupling visual diagnosis from code repair yields an auditable and interpretable evolutionary trace. To substantiate this claim, we provide a detailed taxonomy of the Critic's JSON diagnoses, qualitative examples across different tasks, and an analysis of diagnostic robustness under ambiguous visual evidence.

\subsection{Taxonomy of Diagnoses}
The Critic is prompted to output a structured JSON object containing specific keys that enforce a rigorous diagnostic thought process. This taxonomy ensures that the Actor (and a human auditor) receives standardized, actionable feedback rather than unstructured text. The standard JSON schema includes:

\begin{itemize}[leftmargin=*]
    \item \texttt{failure\_mode}: A high-level categorization of the error (e.g., ``Late Braking'', ``Over-rotation'', ``High Sidelobes'').
    \item \texttt{visual\_evidence}: A description of the specific visual anomalies observed in the Behavioral Evidence (BE) image (e.g., ``The phase portrait shows the pendulum trajectory spiraling outward but failing to close the loop at the upright position'').
    \item \texttt{root\_cause}: The hypothesized algorithmic flaw in the parent policy code that led to the observed visual evidence (e.g., ``The proportional gain $K_p$ is too low to overcome gravity near the top'').
    \item \texttt{suggested\_fix}: A natural language recommendation for the Actor on how to modify the code (e.g., ``Increase $K_p$ or introduce an energy-pumping heuristic when $|\theta| > \pi/2$'').
\end{itemize}

\subsection{Qualitative Examples}

Table~\ref{tab:supp_diagnosis_examples} presents representative examples of the Critic's diagnoses across the evaluated tasks. These examples demonstrate how the Critic grounds its algorithmic hypotheses in concrete visual observations.

\begin{table}[htbp]
  \centering
  \small
  \caption{Representative examples of Critic JSON diagnoses across different tasks.}
  \label{tab:supp_diagnosis_examples}
  \renewcommand{\arraystretch}{1.3}
  \begin{tabular}{p{0.15\linewidth} p{0.8\linewidth}}
    \toprule
    \textbf{Task} & \textbf{Critic Diagnosis JSON (Abridged)} \\
    \midrule
    \textbf{Lunar Lander} & 
    \texttt{\{} \newline
    \texttt{  "failure\_mode": "Premature Engine Cutoff",} \newline
    \texttt{  "visual\_evidence": "The lander's trajectory descends smoothly until $y \approx 0.2$, after which the main engine plume disappears and the lander crashes at high velocity.",} \newline
    \texttt{  "root\_cause": "The altitude threshold for the braking heuristic is set too high, causing the policy to disable the main engine before touchdown.",} \newline
    \texttt{  "suggested\_fix": "Lower the altitude threshold for main engine activation to $y < 0.1$ and ensure the descent rate condition allows for continuous braking."} \newline
    \texttt{\}} \\
    \midrule
    \textbf{Pendulum} & 
    \texttt{\{} \newline
    \texttt{  "failure\_mode": "Insufficient Energy Pumping",} \newline
    \texttt{  "visual\_evidence": "The phase portrait shows the pendulum oscillating back and forth in the lower half-plane, never crossing the $\theta = \pm \pi/2$ horizontal axis.",} \newline
    \texttt{  "root\_cause": "The policy is applying torque proportional to the angle, which fights the natural swing dynamics instead of amplifying them.",} \newline
    \texttt{  "suggested\_fix": "Implement an energy-pumping strategy: apply torque in the direction of the current angular velocity when far from the upright position."} \newline
    \texttt{\}} \\
    \midrule
    \textbf{CCAA Synthesis} & 
    \texttt{\{} \newline
    \texttt{  "failure\_mode": "High Sidelobe Level (SLL)",} \newline
    \texttt{  "visual\_evidence": "The radiation pattern cut shows a prominent first sidelobe at $\theta \approx 45^\circ$ that exceeds the -20 dB target, reaching approximately -15 dB.",} \newline
    \texttt{  "root\_cause": "The amplitude taper drops off too abruptly near the outer rings, creating an aperture illumination that resembles a uniform distribution rather than a smooth taper.",} \newline
    \texttt{  "suggested\_fix": "Smooth the amplitude taper by increasing the excitation of the penultimate ring and slightly decreasing the center element."} \newline
    \texttt{\}} \\
    \bottomrule
  \end{tabular}
\end{table}

\subsection{Robustness and Auditability}

A key advantage of the JSON diagnosis is that it allows human developers to immediately audit the evolutionary process. If a proposed child policy fails, the developer can inspect the JSON to determine whether the failure was due to a \textit{perception error} (the Critic misread the BE) or a \textit{synthesis error} (the Actor wrote buggy code based on a correct diagnosis). 

In our empirical observations, the Critic's diagnostic quality remains robust for well-designed BE representations (like the Lunar Lander trajectories or CCAA radiation cuts). However, we note that diagnostic quality can degrade under ambiguous visual evidence. For example, in the Acrobot task, if the BE renders the double-pendulum links overlapping heavily during a chaotic swing, the Critic occasionally misidentifies the direction of the joint velocities, leading to an incorrect \texttt{root\_cause} hypothesis. Because REFLEX decouples diagnosis from repair, these perception errors are explicitly logged in the \texttt{visual\_evidence} field, making it trivial for researchers to identify the need for a clearer BE rendering function, rather than blindly tuning the LLM prompt.

\section{Experimental Protocol and Hyperparameters}
\label{sec:supp_protocol}

Table~\ref{tab:supp_hparams} lists the settings shared by all formal comparisons in Section~\ref{sec:experiments}. The values are fixed across methods, tasks, and seeds and were not tuned per run.

\begin{table}[h]
  \centering
  \small
  \caption{Hyperparameters and protocol settings for the matched comparisons.}
  \label{tab:supp_hparams}
  \begin{tabular}{p{0.36\linewidth}p{0.56\linewidth}}
    \toprule
    \textbf{Quantity} & \textbf{Value / rule} \\
    \midrule
    \multicolumn{2}{l}{\textit{Search}} \\
    Population size $N$ & 16 \\
    Initialization (control) & Evaluate the task template $p_0$, then fill to $N/2$ with diagnostic (probability 0.6) or text-only mutations of $p_0$; children below $f_{p_0} - \max(1, 0.5|f_{p_0}|)$ are logged but not inserted \\
    Parent selection & Tournament of size 3 \\
    Child insertion & Every valid, non-duplicate child is inserted and the $N$ best are kept; invalid children are logged \\
    UCB1 exploration constant $C$ & 1.2 \\
    UCB1 reward & $\Delta$ clipped to $[0, \Delta_{\max}]$ and divided by $\Delta_{\max}$, the running maximum of positive $\Delta$; failed calls give 0; control tasks also credit ancestors' operators with $0.5^d r$, $d \le 3$ \\
    Forced-burst trigger & 12 consecutive non-improving committed children \\
    Forced burst & One \texttt{m\_inspire} and one \texttt{i1} proposal (control tasks) \\
    Repair and revision & One repair call per crashing child; one revision call when a child of a parent within 5\% of the best regresses by 2--50\% (relative to $\max(1, |f_p|)$) \\
    Skill retrieval & BM25 ($k_1 = 1.5$, $b = 0.75$), top $K = 3$; query = failure modes, root cause, and fix type of $D_p$ \\
    Skill utility & Initial 1; $+\min(\Delta, 1)$ if $\Delta > 0$, else $-\tfrac{1}{2}\min(|\Delta|, 1)$, for retrieved skills the Actor reports using \\
    Skill archiving & $u_i < -1$ after $\ge 3$ retrievals; LLM merge when token Jaccard $> 0.85$ and same target function; lowest utility first above capacity \\
    Skill Memory capacity & 256 skills \\
    Diagnosis reuse & At most 2 Actor samples per Critic diagnosis \\
    \midrule
    \multicolumn{2}{l}{\textit{LLM backend and accounting}} \\
    Model weights (control comparisons) & Qwen/Qwen3-VL-235B-A22B-Instruct for every LLM role (Critic, Actor, repair, abstraction) in REFLEX, Actor-only, and official MLES \\
    Model weights (transfer study) & Qwen3.5-27B \\
    Output cap & 4,096 tokens per call \\
    Sampling workers & 1 \\
    Budget accounting & Every successful or failed LLM generation attempt counts as one call; a diagnostic mutation therefore costs at least two calls (one Critic, one Actor) \\
    \midrule
    \multicolumn{2}{l}{\textit{Evaluation}} \\
    Campaign seeds & 0--9 for all formal comparisons (paired across methods) \\
    Control budgets & 50 and 100 total LLM calls; 20 calls for the tight-budget, memory-object, and transfer studies \\
    MLES baseline & Official implementation at commit \texttt{acda6457}, with native prompts, operators, selection, and duplicate handling \\
    CCAA budget & 240 objective evaluations per run, including REFLEX's inner local-search evaluations \\
    CCAA initialization & Random design vectors, or a shared prior (the same first LLM/analytical designs for every method) \\
    Transfer banks & Top-2 skills by source-task utility, frozen and hash-verified, audited for target-specific identifiers \\
    Statistics & 95\% paired-bootstrap CIs (20,000 resamples for the BE-perturbation study), exact Wilcoxon signed-rank tests, Holm correction within each planned family, Cohen's $d_z$ \\
    \bottomrule
  \end{tabular}
\end{table}

\section{Computational Resources and Execution Time}
\label{sec:supp_compute}

The REFLEX framework is a train-free evolutionary search method. Consequently, it does not require gradient backpropagation or traditional neural network training. The primary computational workload consists of two parts: (1) environment simulation and Behavioral Evidence (BE) rendering, and (2) inference calls to the Large Language Models (LLMs).

All experiments were conducted on a local workstation equipped with 8$\times$ NVIDIA GeForce RTX 3090 GPUs. However, because the LLMs were queried through remote inference endpoints (Qwen/Qwen3-VL-235B-A22B-Instruct for all roles in the matched comparisons; Qwen3.5-27B in the transfer study), the local GPUs were primarily used for parallelizing the Gymnasium environment rollouts, BE image rendering, and CCAA array factor computations.

The wall-clock execution time of a single evolutionary campaign is dominated by inference latency rather than local compute. Throughout the paper, the Critic call and the Actor call of a diagnostic mutation are counted as two separate LLM calls (Table~\ref{tab:supp_hparams}); together they take approximately 3 to 5 seconds on average. For a control-task budget of 100 total LLM calls, a complete evolutionary run takes roughly 10 to 15 minutes. The lightweight nature of the local simulation allows multiple seeds to be evaluated concurrently across the available CPU/GPU resources without significant overhead.

\clearpage
\section*{NeurIPS Paper Checklist}

\begin{enumerate}

\item {\bf Claims}
    \item[] Question: Do the main claims made in the abstract and introduction accurately reflect the paper's contributions and scope?
    \item[] Answer: \answerYes{} 
    \item[] Justification: We clearly state our problem scope and contributions in Sec. \ref{sec:intro}

\item {\bf Limitations}
    \item[] Question: Does the paper discuss the limitations of the work performed by the authors?
    \item[] Answer: \answerYes{} 
    \item[] Justification: Yes, please see Sec. \ref{sec:limitations} for limitations.

\item {\bf Theory assumptions and proofs}
    \item[] Question: For each theoretical result, does the paper provide the full set of assumptions and a complete (and correct) proof?
    \item[] Answer: \answerNA{} 
    \item[] Justification: The paper does not provide any theoretical results.

    \item {\bf Experimental result reproducibility}
    \item[] Question: Does the paper fully disclose all the information needed to reproduce the main experimental results of the paper to the extent that it affects the main claims and/or conclusions of the paper (regardless of whether the code and data are provided or not)?
    \item[] Answer: \answerYes{} 
    \item[] Justification: The method and algorithm are specified in Section~\ref{sec:method} and Algorithm~\ref{alg:reflex}, the experimental protocol in Section~\ref{sec:experiments}, and all hyperparameters in Appendix~\ref{sec:supp_protocol}. The Critic and Actor prompt templates are given in Appendix~\ref{sec:supp_prompts} and example diagnoses in Appendix~\ref{sec:supp_interpretability}; the prompts for the remaining operators are in the released code.

\item {\bf Open access to data and code}
    \item[] Question: Does the paper provide open access to the data and code, with sufficient instructions to faithfully reproduce the main experimental results, as described in supplemental material?
    \item[] Answer: \answerYes{} 
    \item[] Justification: The source code, task evaluators, and commands to reproduce the control and CCAA experiments are publicly available at \url{https://github.com/wangpan-ustc/REFLEX} (linked in the abstract); all tasks are simulators, so no external dataset is required.

\item {\bf Experimental setting/details}
    \item[] Question: Does the paper specify all the training and test details (e.g., data splits, hyperparameters, how they were chosen, type of optimizer) necessary to understand the results?
    \item[] Answer: \answerYes{} 
    \item[] Justification: As a train-free evolutionary framework, our method does not involve traditional neural network training (e.g., data splits or optimizers). However, we specify all evolutionary hyperparameters, LLM-call budgets, and the MLES and Actor-only configurations in Section~\ref{sec:experiments} and Appendix~\ref{sec:supp_protocol} (fixed across methods, tasks, and seeds, not tuned per run); the settings of the CCAA classical optimizers are given in the released code.

\item {\bf Experiment statistical significance}
    \item[] Question: Does the paper report error bars suitably and correctly defined or other appropriate information about the statistical significance of the experiments?
    \item[] Answer: \answerYes{} 
    \item[] Justification: Over 10 paired campaign seeds, we report mean $\pm$ SEM (Tables~\ref{tab:main_results}, \ref{tab:ccaa_baselines}, and \ref{tab:ablation}; exceptions are marked in the captions) and median [IQR] calls or evaluations to target, together with 95\% paired-bootstrap CIs, exact Wilcoxon signed-rank tests with Holm correction, and Cohen's $d_z$ (Section~\ref{sec:experiments}, Appendix~\ref{sec:supp_protocol}).

\item {\bf Experiments compute resources}
    \item[] Question: For each experiment, does the paper provide sufficient information on the computer resources (type of compute workers, memory, time of execution) needed to reproduce the experiments?
    \item[] Answer: \answerYes{} 
    \item[] Justification: Appendix~\ref{sec:supp_compute} reports the hardware, the remote LLM inference setup, and the approximate per-call and per-campaign wall-clock time.
    
\item {\bf Code of ethics}
    \item[] Question: Does the research conducted in the paper conform, in every respect, with the NeurIPS Code of Ethics \url{https://neurips.cc/public/EthicsGuidelines}?
    \item[] Answer: \answerYes{} 
    \item[] Justification: Yes, the research conducted in the paper conforms in every respect with the NeurIPS Code of Ethics.

\item {\bf Broader impacts}
    \item[] Question: Does the paper discuss both potential positive societal impacts and negative societal impacts of the work performed?
    \item[] Answer: \answerNA{} 
    \item[] Justification: NA.
    
\item {\bf Safeguards}
    \item[] Question: Does the paper describe safeguards that have been put in place for responsible release of data or models that have a high risk for misuse (e.g., pre-trained language models, image generators, or scraped datasets)?
    \item[] Answer: \answerNA{} 
    \item[] Justification: NA.

\item {\bf Licenses for existing assets}
    \item[] Question: Are the creators or original owners of assets (e.g., code, data, models), used in the paper, properly credited and are the license and terms of use explicitly mentioned and properly respected?
    \item[] Answer: \answerYes{} 
    \item[] Justification: All third-party assets used in this work are properly cited and credited.

\item {\bf New assets}
    \item[] Question: Are new assets introduced in the paper well documented and is the documentation provided alongside the assets?
    \item[] Answer: \answerNA{} 
    \item[] Justification:  NA.

\item {\bf Crowdsourcing and research with human subjects}
    \item[] Question: For crowdsourcing experiments and research with human subjects, does the paper include the full text of instructions given to participants and screenshots, if applicable, as well as details about compensation (if any)? 
    \item[] Answer: \answerNA{} 
    \item[] Justification: NA.

\item {\bf Institutional review board (IRB) approvals or equivalent for research with human subjects}
    \item[] Question: Does the paper describe potential risks incurred by study participants, whether such risks were disclosed to the subjects, and whether Institutional Review Board (IRB) approvals (or an equivalent approval/review based on the requirements of your country or institution) were obtained?
    \item[] Answer: \answerNA{} 
    \item[] Justification: NA.

\item {\bf Declaration of LLM usage}
    \item[] Question: Does the paper describe the usage of LLMs if it is an important, original, or non-standard component of the core methods in this research? Note that if the LLM is used only for writing, editing, or formatting purposes and does \emph{not} impact the core methodology, scientific rigor, or originality of the research, declaration is not required.
    \item[] Answer: \answerYes{} 
    \item[] Justification: Large Language Models are the core operational components of our REFLEX framework.

\end{enumerate}

\end{document}